\documentclass[sigconf]{acmart}
\usepackage{multirow}
\usepackage{arydshln}

\usepackage{graphicx}
\usepackage{arydshln}
\AtBeginDocument{%
  }

\usepackage{latexsym}
\usepackage{graphicx}
\usepackage{enumitem}
\usepackage{pgfplots}
\usepackage{amsmath}
\usepackage{xcolor,colortbl}

\let\tb=\textbf

\begin{document}

\title{Efficient and Accurate Scene Text Recognition with Cascaded-Transformers}

\author{Savas Ozkan}
\email{savas.ozkan@samsung.com}
\affiliation{%
  \institution{Samsung Research}
  \country{United Kingdom}
}

\author{Andrea Maracani}
\affiliation{%
  \institution{Samsung Research}
  \country{United Kingdom}
}

\author{Hyowon Kim}
\affiliation{%
  \institution{Samsung Electronics}
  \country{South Korea}}
  
\author{Sijun Cho}
\affiliation{%
  \institution{Samsung Electronics}
  \country{South Korea}}

\author{Eunchung Noh}
\affiliation{%
  \institution{Samsung Electronics}
  \country{South Korea}}

\author{Jeongwon Min}
\affiliation{%
  \institution{Samsung Electronics}
  \country{South Korea}}

\author{Jung Min Cho}
\affiliation{%
  \institution{Samsung Electronics}
  \country{South Korea}}
  
\author{Mete Ozay}
\affiliation{%
  \institution{Samsung Research}
  \country{United Kingdom}
}

\renewcommand{\shortauthors}{Ozkan et al.}

\begin{abstract}

In recent years, vision transformers with text decoder have demonstrated remarkable performance on Scene Text Recognition (STR) due to their ability to capture long-range dependencies and contextual relationships with high learning capacity. However, the computational and memory demands of these models are significant, limiting their deployment in resource-constrained applications. To address this challenge, we propose an efficient and accurate STR system. Specifically, we focus on improving the efficiency of encoder models by introducing a cascaded-transformers structure. This structure progressively reduces the vision token size  during the encoding step, effectively eliminating redundant tokens and reducing computational cost. Our experimental results confirm that our STR system achieves comparable performance to state-of-the-art baselines while substantially decreasing computational requirements. In particular, for large-models, the accuracy remains same, $92.77 \rightarrow 92.68$, while computational complexity is almost halved with our structure.

\end{abstract}




\maketitle

\section{Introduction}

Scene Text Recognition (STR) is a complex task that involves interpreting and extracting textual content from natural scenes, and converting it into digital character or word sequences. This process enables to estimate high-level semantics, which is crucial for scene understanding and perception. The ability to accurately recognize and extract text from scenes has numerous applications and use cases, including document scanning, navigation, robotics, and many other products and services.

Compared to traditional Optical Character Recognition (OCR) techniques~\cite{islam2017survey}, which typically operate on scanned or digital documents, STR settings pose significant challenges. The presence of variations in text deformation, occlusions, and cluttered backgrounds can all undermine the performance of STR systems. 
%For instance, text in natural scenes may be distorted, rotated, or partially obscured, making it difficult for algorithms to accurately detect and recognize. Additionally, the wide range of fonts, styles, and sizes used in natural scenes can further complicate the text recognition process. 
To address these challenges, STR systems must be designed to be robust and adaptable to various adverse conditions and text formats. This requires the development of advanced algorithms and techniques that can effectively handle the complexities of natural scenes.
%, including varying lighting conditions, noise, and other forms of degradation. Furthermore, STR systems must be able to generalize well to new and unseen data.

Recent breakthroughs in deep learning and computer vision have revolutionized the Scene Text Recognition (STR) domain, enabling the development of more accurate and effective systems. In particular, STR systems have widely adopted deep neural networks, such as convolutional neural networks (CNNs)~\cite{wang2012end}, recurrent neural networks (RNNs)~\cite{su2015accurate} and Transformers~\cite{atienza2021vision}, which are trained on large datasets to handle the complexities of text recognition in natural scenes. These deep networks are designed to learn robust features from large training samples, allowing them to effectively tackle challenging conditions.

The unprecedented success of deep learning models in STR can be attributed to the fact that larger datasets and deeper networks tend to yield better performance. %As a result, researchers have been driven to collect and annotate large-scale datasets, and to design increasingly complex architectures to improve accuracy. This has led to notable improvements in recognition accuracy, with many state-of-the-art models achieving impressive performance on benchmark datasets. 
One promising research direction in the field of Scene Text Recognition (STR) focuses on utilizing deeper networks based on transformer architectures to effectively capture high-level details of samples in adverse scenarios~\cite{rang2023large}. These transformer-based models have been shown to generalize model predictions better than smaller networks, resulting in lower error rates and improved overall performance.
%The key insight behind this approach is that increasing the capacity of the encoder models of the network, such as the number of layers or the number of attention heads, has a more significant positive impact on performance compared to increasing the capacity of the decoder models. 

In particular, the use of transformer-based architectures allows for the effective capture of long-range dependencies and contextual relationships within the input data, which is critical for accurately recognizing text in complex scenes. % By increasing the model's capacity, the network is able to learn more abstract representations of the input data, leading to improved performance. The results of this research direction have been impressive, with transformer-based models obtaining state-of-the-art performance on benchmark datasets. However, this improved performance comes at a cost.
One significant bottleneck of these deep learning models is that they require substantial computational and memory resources to operate effectively. As the models grow in complexity and size, their computational and memory demands increase exponentially, making them less practical for deployment on edge devices with limited resources. The high computational requirements of these models can lead to increased power consumption, heat generation, and latency, which can be detrimental to the overall performance and user experience. %Therefore, while the use of transformer-based architectures has shown great promise in STR, it is essential to explore ways to make these models more efficient and compact without sacrificing performance.

\begin{figure}[t]
\centering
\includegraphics[width=0.45\textwidth]{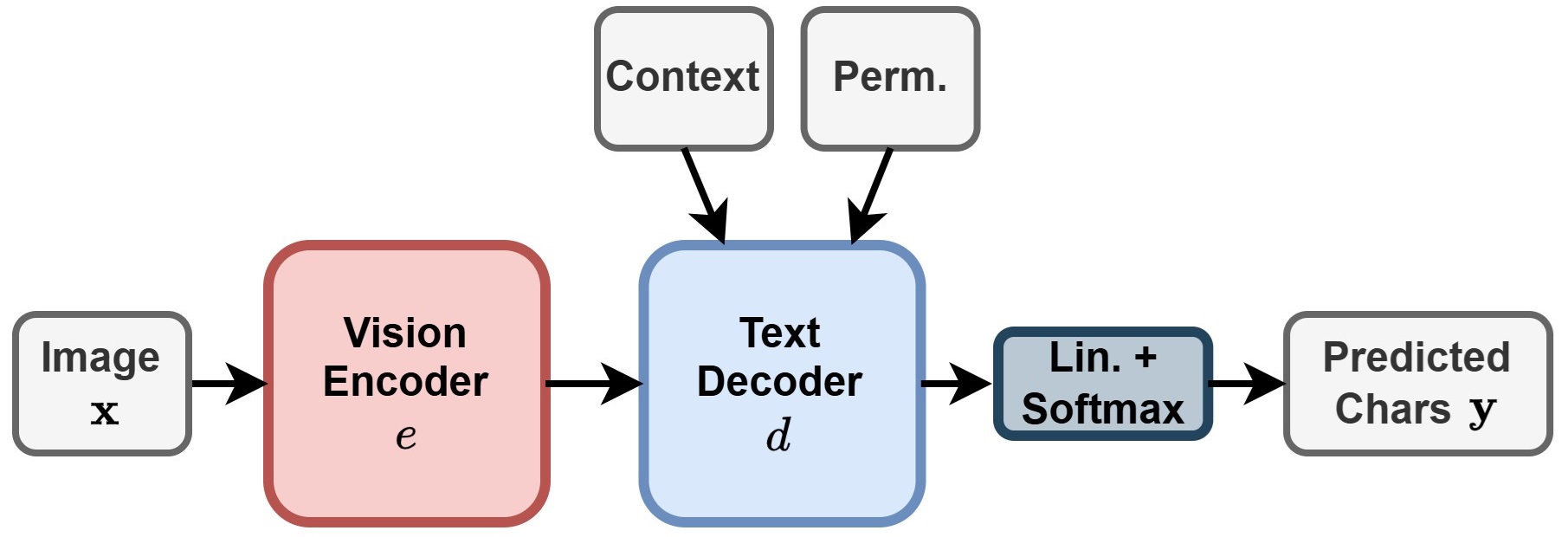}
\vspace{-0.8em}
\caption{\textbf{The overall architecture of our STR system}. Our system consists of encoder $e$ and decoder $d$ models and predicts text characters $\mathbf{y}$ from an input image $\mathbf{x}$.}
\vspace{-0.8em}
\label{fig:architecture}
\end{figure}

Another notable work presents a new paradigm for the STR domain that a single vision token can be sufficient to accurately decode the text, eliminating the need for sequential tokens~\cite{xu2024ote}. This proves two significant advantages for STR domain: 1) it avoids the alignment issue between vision tokens and character embeddings during decoding step. 2) it provides computational efficiency during the decoding step by allowing for adjustable vision tokens. By adjusting the number of vision tokens, the model can be optimized for computational efficiency, making it more suitable for resource-constrained devices. This is particularly important for applications where computational resources are limited, such as mobile devices or edge devices. However, it is essential to note that the computational efficiency gained by this approach may be relatively small compared to the overall computational demands of the model.
%As discussed in the context of model scaling, the majority of the computational demands in STR models are typically caused during the encoding step, where the vision tokens are extracted and processed. 

Moreover, there have been several attempts~\cite{zhang2020autostr} to search for the optimal Scene Text Recognition (STR) encoder by leveraging the concept of neural architecture search (NAS)~\cite{elsken2019neural}. The work in this area has shown that the capacity of encoders can be effectively tuned by considering the constraints imposed by the data and model complexity. This approach enables the automatic design of encoders that are tailored to the specific requirements of the STR task, potentially leading to improved performance and efficiency.
%However, the usability of this approach has only been validated on convolution-based models, which may limit its applicability to other types of state-of-the-art models, such as transformer-based architectures. 
Ideally, the goal should be to develop STR models that are not only highly accurate but also efficient, scalable, and practical for real-world application and use-cases.

In this paper, we propose a STR system that aims to increase the efficiency of STR models in terms of computational complexity with a negligible accuracy drop. For this purpose, the model complexity is decreased by hierarchically selecting a subset of vision tokens and learning to discard redundant vision tokens during training step. The motivation of our system is that text content in provided images is typically sparse (i.e., a significant portion of the content covers background rather than text information), therefore there are redundant vision tokens that can be discarded in the early levels of encoder computations. To this end, the efficiency of STR system can be improved. Our contributions can be summarized as follows:
\begin{itemize}[leftmargin=*]
    \item We propose a novel cascaded-transformers structure that replaces the standard vision-transformer encoder for computer vision problems. This structure consists of a series of sub-transformers that selectively retain the most important vision tokens at each level, reducing computational requirements while preserving essential information. Details of cascaded transformers are presented in Section~\ref{subsection:ect}.
    
    \item We demonstrate its superiority on STR domain that the proposed model improves the efficiency while achieving compatible results with state-of-the-art baselines.
    
    \item Moreover, we implement a custom software setup to test our STR system on real-world images. Particularly, our system achieves high true prediction rate under highly distorted, occluded and rotated text scenarios.
    
\end{itemize}

This paper is organized as follows: first, we will formulate the STR problem and then we will explain the details of our STR system with novel cascaded transformers structure. Later, we will present our experimental results and conclude our paper.

\section{Problem Formulation}

Given a natural scene image containing text, the task of scene text recognition (STR) aims to transcribe each individual character within the text present in the image. 

Formally, let $\mathcal{X} \subseteq \bigcup_{H,W} \mathbb{R}^{H \times W \times 3}$ be the space of text images and $\mathcal{Y} := \{ (y_0, \ldots, y_{L-1}): L \in \mathbb{N}^+, y_i \in \mathcal{C}\}$ be the space of sequences of characters of strictly positive length from a character set $\mathcal{C}$. We assume the existence of an underlying data distribution $\mathcal{D}$ over the set $\mathcal{X} \times \mathcal{Y}$, and we define a loss function $\ell: \mathcal{Y} \times \mathcal{Y} \to \mathbb{R}$ to measure the difference between two sequences. The objective of STR is to find a prediction function $f: \mathcal{X} \to \mathcal{Y}$ that minimizes the expected loss, $\mathbb{E}_{(\mathbf{x}, \mathbf{y}) \sim \mathcal{D}}[\ell(f(\mathbf{x}),\mathbf{y})]$. 
 The function $f$ is implemented by a deep network $\phi$ that is parameterized by $\theta$ and trained on a dataset $\mathcal{S} = \{(\mathbf{x}^i, \mathbf{y}^i)\}_{i=1}^M$ that contains $M$ i.i.d. samples from $\mathcal{D}$.

\begin{figure*}[t]
\centering
\includegraphics[width=0.75\textwidth]{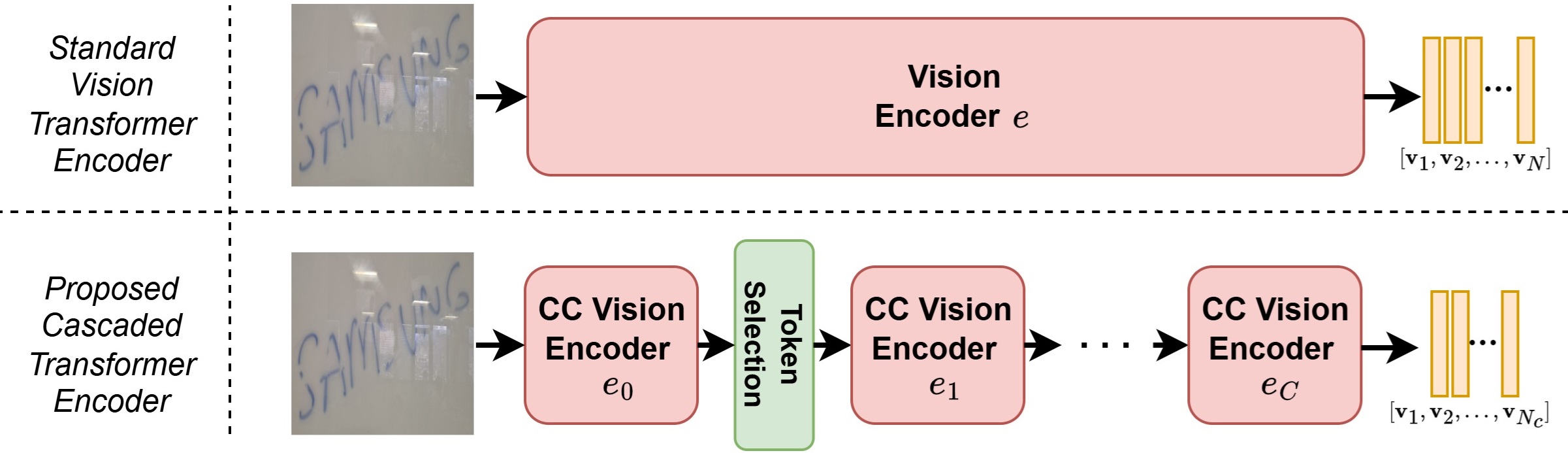}
\vspace{-0.8em}
\caption{\textbf{The flows of standard vision transformer and proposed cascaded transformers}. Both models take an input image $\mathbf{x}$ and output vision tokens $\mathbf{v}$. The key component of our proposed model is a series of cascaded transformers, which reduces the number of vision tokens at each level by selectively retaining a subset of the most relevant ones for scene text recognition (i.e., $N < N_C$). Details are explain in Section~\ref{subsection:ect}.}
\vspace{-0.8em}
\label{fig:flow}
\end{figure*}

\section{Our Scene Text Recognition System}

In our system, we employ an encoder-decoder architecture. The overall architecture is illustrated in Figure~\ref{fig:architecture}. The input image $\mathbf{x} \in \mathcal{X}$ is first projected into a vision tokens $\mathbf{v} \in \mathcal{V}$ comprising a sequence of $N$ $K$-dimensional vision tokens via an encoder function $e: \mathcal{X} \to \mathcal{V} \subset \mathbb{R}^{N \times K}$. Later, the vision tokens $\mathbf{v}$ is mapped to a character sequence $\mathbf{y} \in \mathcal{Y}$ using a decoder function $d: \mathcal{V} \to \mathcal{Y} \subset \mathbb{R}^{L}$. To this end, our network is decomposed as $\phi(x)=d \circ e(x)$ and both models (i.e., encoder and decoder models) are based on transformers. Ultimately, our objective is to improve the efficiency of encoder $\mathbf{e}$ while preserving the same representation capacity of latent vision space $\mathcal{V}$ for high true prediction rates.

\vspace{0.2em}
\noindent
\textbf{Decoder Model $d$:} A conditional language model is utilized. Specifically, similar to prior works~\cite{bautista2022scene}, we leverage a Permuted-Language Decoder (PLD), which introduces a positional query stream separated by the key-value stream enabling to decode characters in different orders and incorporating Permutation Language Modeling (PLM) during training. Formally, let $\mathbf{z} = (z_1, z_2, \ldots, z_{L-1})$ be a permutation establishing an ordering among the characters and let $y_{\mathbf{z}<t}$ be the context up to current step $t$ (i.e., already predicted characters), following the order specified by $\mathbf{z}$. Then, the decoder $d(\mathbf{z}_t, \mathbf{v}, \mathbf{y}_{\mathbf{z}<t})$ outputs a probability distribution over the character set $\mathcal{C}$ for predicting character at position $z_t$. The PLD is implemented as a block that contains two cross-attention layers and an MLP. Positional queries are used in the query stream of the decoder to specify the position to predict according to permutation $\mathbf{z}$. In the key-value stream, context is introduced in the first cross-attention layer, which is masked during training to simulate the $\mathbf{z}$ order without actually permuting the characters. Vision tokens $\mathbf{v}$ are then incorporated in the second cross-attention layer. The capacity of decoder model $d$ can be increased using multiple cascaded blocks. We observe that when context and vision tokens are not updated in the multiple block scenario, a clear accuracy improvement can be observed. Indeed, increasing block size negatively affects the computational and memory demands.

\begin{table}[t]
  \centering
  
  %\settowidth\rotheadsize{\tb{ENCODER}}
  \addtolength{\tabcolsep}{-0.3em}
  \resizebox{0.9\linewidth}{!}{
  \begin{tabular}{clcccc|ccc}
    \toprule
    & & & & & &  \multicolumn{3}{c}{\tb{GFLOPs}} \\
    & & \tb{Blocks} & \tb{Dim} & \tb{Heads} & \tb{Params} & $P=8$ & $P=16$ & $P=32$ \\
    \midrule
    \multirow{3}{*}{\rotatebox[origin=c]{90}{\scriptsize \tb{ENC}}} 
    & $e$-tiny  & 12 & 192 & 3  & 5.5 M  & 8.8   & 2.2  & 0.5 \\
    & $e$-small & 12 & 384 & 6  & 21.7 M & 34.4  & 8.6  & 2.1 \\
    & $e$-base  & 12 & 768 & 12 & 85.8 M & 135.6 & 33.9 & 8.5 \\
       
    \midrule
    \multirow{2}{*}{\rotatebox[origin=c]{90}{\scriptsize \tb{DEC}}} 
    & $d$-small & 1 & 768 & 12 & 9.6 M & 8.7 & 3.5 & 2.2 \\
    & $d$-base & 2 & 768 & 12 & 19.1 M & 17.5 & 7.0 & 4.4 \\
    \bottomrule
  \end{tabular} }
  \caption{\textbf{Details of encoder (ENC) and decoder (DEC) used in our STR system.} In particular, the impact of patch size $P$ on the overall computational complexity is demonstrated.}
  \vspace{-0.8em}
  \label{tab:arch_comp}
\end{table}

\vspace{0.2em}
\noindent
\textbf{Encoder Model $e$:} A vision transformer architecture that is pre-trained on ImageNet21k~\cite{deng2009imagenet}, is utilized. Similar to the original pipeline~\cite{alexey2020image}, an input image $\mathbf{x}$ is converted into a sequence of 2D patches whose resolution is $(P\times P)$. Hence, $N=HW/P^2$ number of patches is extracted per image. Ultimately, when $N$ is high, computations required in self-attentions become high. On the other hand, when $N$ is low, the representation capacity of transformers is expected to be reduced. Later, these patches are projected to patch embeddings using a linear layer and 1D position embeddings are added to retain positional information. Later, $T$ number of multi-headed self-attentions with layer norm is used to compute vision tokens $\mathbf{v}$. In this paper, our objective is to improve the efficiency of encoder $e$. Unlike the standard architecture, we selectively reduce the number of patches $N$ while computing vision tokens $\mathbf{v}$. To this end, the computational demands can be significantly decreased. Details of our contribution will be explained in~\ref{subsection:ect}.

To optimize our system (i.e., both encoder and decoder models) in the training step, we minimize the next-permuted-token prediction cross-entropy loss over random permutations as follow:

\begin{equation}
\min_{\theta} \mathbb{E}_{\substack{{(\mathbf{x}, \mathbf{y}) \sim \mathcal{D}} \\ {\mathbf{z} \sim \mathcal{Z}}}} \left[  \sum_{t=1}^{L} -\mathbf{1}(\mathbf{y}_{\mathbf{z}_t}) \cdot \log d(\mathbf{z}_t, \mathbf{v}, \mathbf{y}_{\mathbf{z}<t}) \right]
\label{eq:PLM_loglikelihood}
\end{equation}

\noindent
Here, the parameters of network $\theta$ are optimized by minimizing this loss function.

\subsection{Efficient Cascaded-Transformers Structure}
\label{subsection:ect}

In our system, we conduct a comprehensive analysis of various encoder and decoder models with different capacities, focusing on their memory and computational complexities. Our analysis reveals that the majority of the computational demands is incurred during the computation of the vision tokens $\mathbf{v}$ from the input image $\mathbf{x}$. This suggests that the encoding process is the primary bottleneck in terms of computational efficiency.

\begin{table}[t]
  \centering
  
  %\settowidth\rotheadsize{\tb{ENCODER}}
  \addtolength{\tabcolsep}{-0.3em}
  \resizebox{0.7\linewidth}{!}{
  \begin{tabular}{c|ccc}
    \toprule
    & \multicolumn{3}{c}{\tb{Acc (\%)}} \\
    & $P=8$ & $P=16$ & $P=32$ \\
    \midrule
    $e$-tiny + $d$-tiny  & 91.71 & 91.01 & 88.53 \\
    $e$-small + $d$-tiny & 92.35 & 92.24 & 90.21 \\
    $e$-base + $d$-tiny  & 93.34 & 92.77 & 91.71 \\
    \bottomrule
  \end{tabular} }
  \caption{\textbf{Impact of patch resolution $P$ on the word accuracy.} Standard encoder and decoder models are utilized and the average word accuracy is reported for different patch resolution $P$.}
  \vspace{-0.8em}
  \label{tab:arch_acc}
\end{table}

To provide a detailed understanding of these models, we present their hyperparameters related to transformers, parameter size, and GFLOPs in Table~\ref{tab:arch_comp}. The table illustrates the impact of patch resolution $P$ (intuitively patch size $N$ as well) on the overall computational complexity. Also, we report the impact of patch resolution $P$ to the word accuracy in Table~\ref{tab:arch_acc}. It is evident that the patch size is a crucial factor affecting the efficiency and accuracy of the STR system. Specifically, increasing the patch size leads to higher model complexity, while reducing its size adversely affects the word accuracy. This trade-off between patch size and word accuracy highlights the need for a balanced approach to achieve efficient and accurate STR system.

Based on this observation, we propose a novel cascaded transformer-based structure for the encoder $e$. Our approach involves dividing the transformer model into several sub-models, each of which processes a subset of vision tokens in a hierarchical manner. At the output of each sub-model, a selection process is applied to choose a subset of the most relevant vision tokens, which are then passed to the next sub-model. By incrementally refining the representation of the vision tokens through this cascaded process, we aim to reduce the computational complexity of the STR system while preserving its accuracy. The selection of vision tokens at each level allows us to retain the most important information while discarding less relevant tokens, thereby decreasing the dimensionality of the data and the computational requirements. The architecture is depicted in Figure~\ref{fig:flow}, which illustrates the hierarchical processing of vision tokens through the cascaded sub-models.

In our system, instead of using a single encoder model $e$, we employ a cascaded mechanism consisting of a series of sub-models, denoted as $[e_1, e_2,..., e_C]$, where $C$ represents the number of cascaded models. This design allows for the adjustment of the number of vision tokens at each level, enabling a more efficient processing pipeline. Formally, each sub-model $e_i$ takes the previously processed vision token $\mathbf{v_{i-1}} \in \mathbb{R}^{N_{i-1} \times K}$ as input and computes the current vision tokens $\mathbf{v_{i}} \in \mathbb{R}^{N_i \times K}$ by $\mathbf{v_i}=e_i(\mathbf{v_{i-1}})$. Notably, the number of vision tokens is progressively reduced at each level as $N_i<N_{i-1}$ to improve the STR efficiency. This reduction in vision tokens is achieved through the selection of the most relevant information, allowing our system to focus on the most critical aspects of the input data (i.e., text content rather than background). To this end, our encoding step is formulated as follow: 

\begin{equation}
e(x)=e_C \circ ... \circ e_2 \circ e_1(x)
\end{equation}

In our system, we deliberately maintain the total size of the parameters unchanged as $\theta_e = \theta_{e_1} \cup \theta_{e_2} \cup... \cup \theta_{e_C}$. This ensures that the overall parameter size remains constant, while the model complexity is optimized for efficiency. 

For the selection of vision tokens, we investigate various selection schemes, including 1D local average pooling, 1D local max pooling, 2D convolution with strides and selecting the first $N_i$ vision tokens (including the CLS token). Empirically, we find that selecting the first $N_i$ vision tokens yields the best accuracy among the considered schemes.

\begin{figure}[t]
\centering
\includegraphics[width=0.4\textwidth]{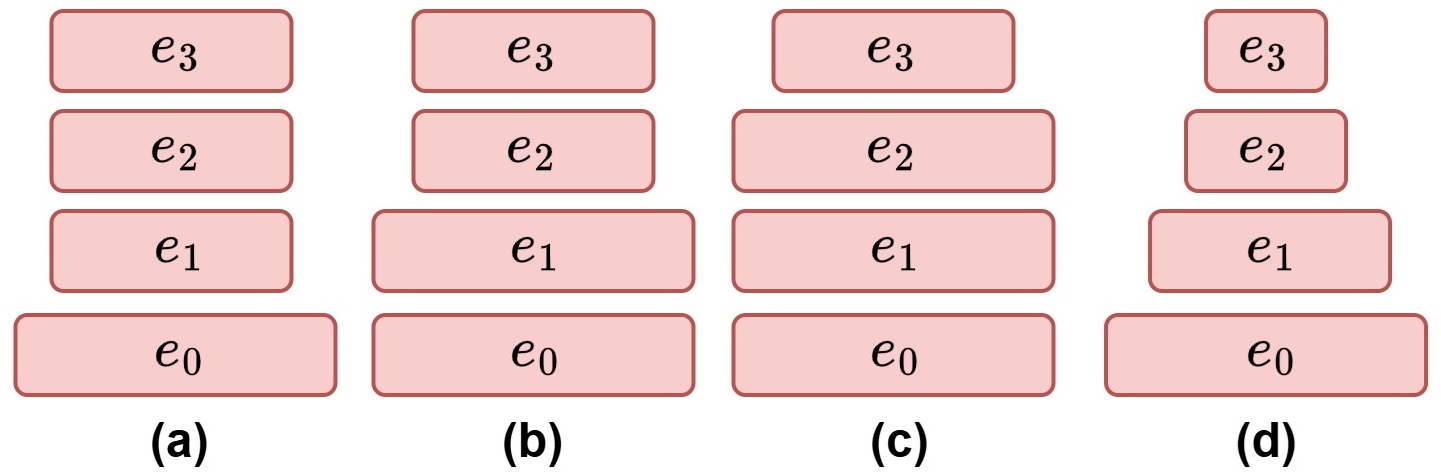}
\vspace{-0.8em}
\caption{\textbf{Four different reduction styles for cascaded transformers: (a) early-level reduction (b) mid-level reduction (c) late-level reduction, and (d) multi-level reduction}. }
\vspace{-0.8em}
\label{fig:cascaded_app}
\end{figure}

Moreover, we examine the impact of the way of reducing the number of vision tokens in the cascaded structure. Specifically, we analyze whether it is more effective to reduce the number of vision tokens in the early, middle, or late levels of the cascaded structure. To this end, we conduct experiments with four different reduction styles as illustrated in Figure~\ref{fig:cascaded_app}: (a) early-level reduction, (b) mid-level reduction, (c) late-level reduction, and (d) multi-level reduction. The early and mid-level reductions involve starting to reduce the number of vision tokens either in the initial and mid levels of the cascade, while the late-level reduction delays the reduction until the later levels. On the other hand, the multi-level reduction reduces the number of vision tokens uniformly across all levels. Our experiments aim to analyze which approach leads to the best trade-off between accuracy and efficiency.

%In our system, we consider different encoder and decoder models with different capacities by comparing their memory and computational complexities. Throughout our analysis, we identify that the majority of the computational demands are caused while computing the latent representation $\mathbf{v}$ from the input image $\mathbf{x}$. Specifically, the details of these models in terms of hyperparameters related to transformers, parameter size and GLOPs are illustrated in Table ?. In the table, we also show that how patch size $N$ impacts the overall computational complexity. It is clear that patch size is one of the factors that affects the efficiency of STR system. Precisely, higher patch size introduces higher model complexity. However, when we reduces its size, the word accuracy is adversely affected. 

%For this purpose, we introduce a cascaded model for encoder model based on transformers. The encoder model is splitted into several sub-models and at the output of each sub-model, a subset of vision tokens is hierarchically selected to transfer them to the next sub-model. The flow is illustrated in Figure ?.

\section{Experiments}

\subsection{Datasets and Metric}

Training large models for the STR domain has historically been a challenging task due to the limited availability of real-world labeled data. However, with the increasing accessibility of real data with corresponding labels, it is possible to train models on both real and synthetic datasets. The training scheme that uses both real and synthetic datasets has been shown to be sample-efficient and can lead to better accuracy, as the models can learn from the diversity of real-world and synthetic data.

\vspace{0.2em}
\noindent
\textbf{Real Datasets (R) ($3.3$M):} To train our STR model with real data, we use  COCO-Text~\cite{veit2016coco}, RCTW17~\cite{shi2017icdar2017}, Uber-Text~\cite{zhang2017uber}, ArT~\cite{chng2019icdar2019}, LSVT~\cite{sun2019icdar}, MLT19~\cite{nayef2019icdar2019}, TextOCR~\cite{singh2021textocr}, ReCTR~\cite{zhang2019icdar} and OpenVINO~\cite{krylov2021open}. There are several challenging cases such as low-resolution, occluded, curved and rotated text in samples. The detail analysis of these datasets is presented in~\cite{jiang2023revisiting}. 

\vspace{0.2em}
\noindent
\textbf{Benchmark Datasets (B) ($15$K):} The benchmark datasets used in the training step are IIIT5K~\cite{mishra2012scene}, Street View Text (SVT)~\cite{wang2011end}, ICDAR13~\cite{karatzas2013icdar} and ICDAR15~\cite{karatzas2015icdar}.

\vspace{0.2em}
\noindent
\textbf{Union14M Dataset (U) ($3$M):} We use another real and large dataset to test the impact of data scaling that is collected from 15 publicly available sub-datasets and contains approximately $3$M of labeled data~\cite{jiang2023revisiting}.

\vspace{0.2em}
\noindent
\textbf{Synthetic Datasets (S) ($6$M):} A subset of MJSynth (MJ)~\cite{jaderberg2014synthetic} and SynthText (ST)~\cite{gupta2016synthetic} is sampled to analyze the scalability of our STR model on synthetic datasets. Experimentally, we observe that a data ratio of $1:0.5$ between real-synthetic datasets obtains the best performance.

\vspace{0.2em}
\noindent
\textbf{Test Datasets:} For evaluation benchmarks, we use IIIT5K~\cite{mishra2012scene}, CUTE-80 (C80)~\cite{risnumawan2014robust}, Street View Text (SVT)~\cite{wang2011end}, Street View Text-Perspective (SVT-P)~\cite{phan2013recognizing}, ICDAR13 (IC13)~\cite{karatzas2013icdar} and ICDAR15 (IC15)~\cite{karatzas2015icdar}. Furthermore, we report performance on HOST and WOST~\cite{wang2021two} datasets that mostly have visually occluded examples. Lastly, more recent benchmarks, COCO-Text~\cite{veit2016coco}, ArT~\cite{chng2019icdar2019} and Uber-Text~\cite{zhang2017uber}, are also utilized.

\vspace{0.2em}
\noindent
\textbf{Evaluation Metric:} To evaluate our system, we follow the same procedure as presented in previous works~\cite{zhao2023clip4str, rang2023large}. Hence, word accuracy metric is used where a predicted character sequence is accepted as correct if all characters match with the ground truth label. Average word accuracy across all 11 test benchmarks is reported. Also, floating-point operations (GFLOPs) are reported using the implementation\footnote{https://github.com/MrYxJ/calculate-flops.pytorch}.

\subsection{Implementation Details}

Input images are augmented by adding random Gaussian blur and noise to provide additional robustness to our system for adverse conditions. Furthermore, all images are resized to $224 \times 224$ and patch size $P$ is set to $16$ by default. Therefore, $196$ vision tokens are initially computed per image. Maximum character length $L$ is $25$. During training, a set $\mathcal{C}$ of $94$ characters is used.

For model optimization, we use AdamW~\cite{loshchilov2017decoupled} with a learning rate of $0.001$. The training process involves a batch size of $1024$ and a total of $10$ epochs. Also, $\beta_1$ and $\beta_2$ are set to $0.9$ and $0.95$, respectively. To train Permutation Language Model (PLM), $4$ random permutations are sampled in each iteration.

\subsection{Experimental Results}

In our experimental results, we denote the standard vision transformer encoder by $e$-model$_{id}$, where model$_{id}$ refers to the model capacity, which can be one of three variants: \textit{tiny}, \textit{small}, or \textit{base} (see Table~\ref{tab:arch_comp} for details). On the other hand, our cascaded-transformers encoder is represented by $e$-cc($i:j:...:k$)-model$_{id}$ where $i$, $j$ and $k$ specify the block size of each sub-transformer. The notation $:$ indicates that vision token size is halved after the current transformer-block in our implementation.

\begin{table}
  \centering
  
  %\makegapedcells
  %\settowidth\rotheadsize{\tb{ENCODER}}
  \addtolength{\tabcolsep}{-0.3em}
  \resizebox{0.8\linewidth}{!}{
  \begin{tabular}{c|cccc}
    \toprule
    & \multicolumn{2}{c}{$d$-tiny} & \multicolumn{2}{c}{$d$-small} \\
    & \textbf{Acc} & \textbf{GFLOPs} & \textbf{Acc} & \textbf{GFLOPs} \\
    \midrule
    $e$-tiny         & 91.01 & 5.7 & 91.77 & 9.2 \\
    $e$-cc(3:9)-tiny & 90.49 & 4.0 & 91.33 & 5.9 \\
    $e$-cc(6:6)-tiny & 90.84 & 4.2 & 91.36 & 6.1 \\
    $e$-cc(9:3)-tiny & 90.56 & 4.5 & 91.79 & 6.4 \\
    \midrule
    $e$-small         & 92.24 & 12.1 & 92.49 & 15.6 \\
    $e$-cc(3:9)-small & 91.90 & 7.5  & 92.43 & 9.7 \\
    $e$-cc(6:6)-small & 91.94 & 9.0  & 92.46 & 10.9 \\
    $e$-cc(9:3)-small & 91.89 & 9.6  & 92.56 & 11.5 \\
    \midrule
    $e$-base         & 92.77 & 37.4 & 93.21 & 40.9 \\
    $e$-cc(3:9)-base & 92.68 & 23.1 & 93.09 & 25.1 \\
    $e$-cc(6:6)-base & 92.98 & 28.0 & 93.39 & 29.9 \\
    $e$-cc(9:3)-base & 92.72 & 31.6 & 93.26 & 33.6 \\
    \bottomrule
  \end{tabular}}
  \caption{\textbf{Results for two-level cascaded transformers.} Average word accuracy (\%) on the test benchmarks is reported with the GFLOPs of overall STR system.}
  \vspace{-0.8em}
  \label{tab:results_two}
\end{table}

\subsubsection{Two-Level Cascaded Transformers}

In this section, we analyze the impact of two-level cascaded transformers in terms of prediction word accuracy and model complexity. Specifically, we divide the encoder model into two cascaded transformers where $C=2$. In table~\ref{tab:results_two}, we present the experimental results for different encoder and decoder model combinations. Intuitively, $e$-cc(3:9)-model$_{id}$, $e$-cc(6:6)-model$_{id}$ and $e$-cc(9:3)-model$_{id}$ represent different reduction styles as early-level, mid-level and late-level, respectively. As expected, $e$-cc(3:9)-model$_{id}$ reduces the model complexity most compared to the standard encoder. In term of accuracy, $e$-cc(6:6)-model$_{id}$ consistently obtains the best results among all configurations. For large-models (i.e., \textit{base}), the cascaded-transformers model even outperforms the standard transformer.

\begin{table}
  \centering
  
  %\settowidth\rotheadsize{\tb{ENCODER}}
  \addtolength{\tabcolsep}{-0.3em}
  \resizebox{0.8\linewidth}{!}{
  \begin{tabular}{c|cccc}
    \toprule
    & \multicolumn{2}{c}{$d$-tiny} & \multicolumn{2}{c}{$d$-small} \\
    & \textbf{Acc} & \textbf{GFLOPs} & \textbf{Acc} & \textbf{GFLOPs} \\
    \midrule
    $e$-tiny             & 91.01 & 5.7 & 91.77 & 9.2 \\
    $e$-cc(4:4:4)-tiny   & 90.14 & 3.3 & 91.17 & 5.5 \\
    %$e$-cc(3:6:3)-tiny   & ??.?? & ??.?? & ??.?? & ??.?? \\
    $e$-cc(3:3:3:3)-tiny & 89.18 & 2.9 & 90.35 & 4.9 \\
    \midrule
    $e$-small             & 92.24 & 12.1 & 92.49 & 15.6 \\
    $e$-cc(4:4:4)-small   & 91.41 & 7.1 & 92.38 & 9.3 \\
    %$e$-cc(3:6:3)-small   & ??.?? & ??.?? & ??.?? & ??.?? \\
    $e$-cc(3:3:3:3)-small & 91.63 & 5.9 & 92.20 & 7.9 \\
    \midrule
    $e$-base             & 92.77 & 37.4 & 93.21 & 40.9 \\
    $e$-cc(4:4:4)-base   & 92.42 & 24.2 & 92.91 & 19.8 \\
    %$e$-cc(3:6:3)-base   & ??.?? & ??.?? & ??.?? & ??.?? \\
    $e$-cc(3:3:3:3)-base & 92.16 & 21.9 & 92.78 & 17.8 \\
    \bottomrule
  \end{tabular} }
  \caption{Results for multi-level cascaded transformers. Average word accuracy (\%) on the test benchmarks is reported with the GFLOPs of overall STR system.}
  \vspace{-0.8em}
  \label{tab:results_mult}
\end{table}

\begin{figure*}[t]
\centering
\includegraphics[width=0.75\textwidth]{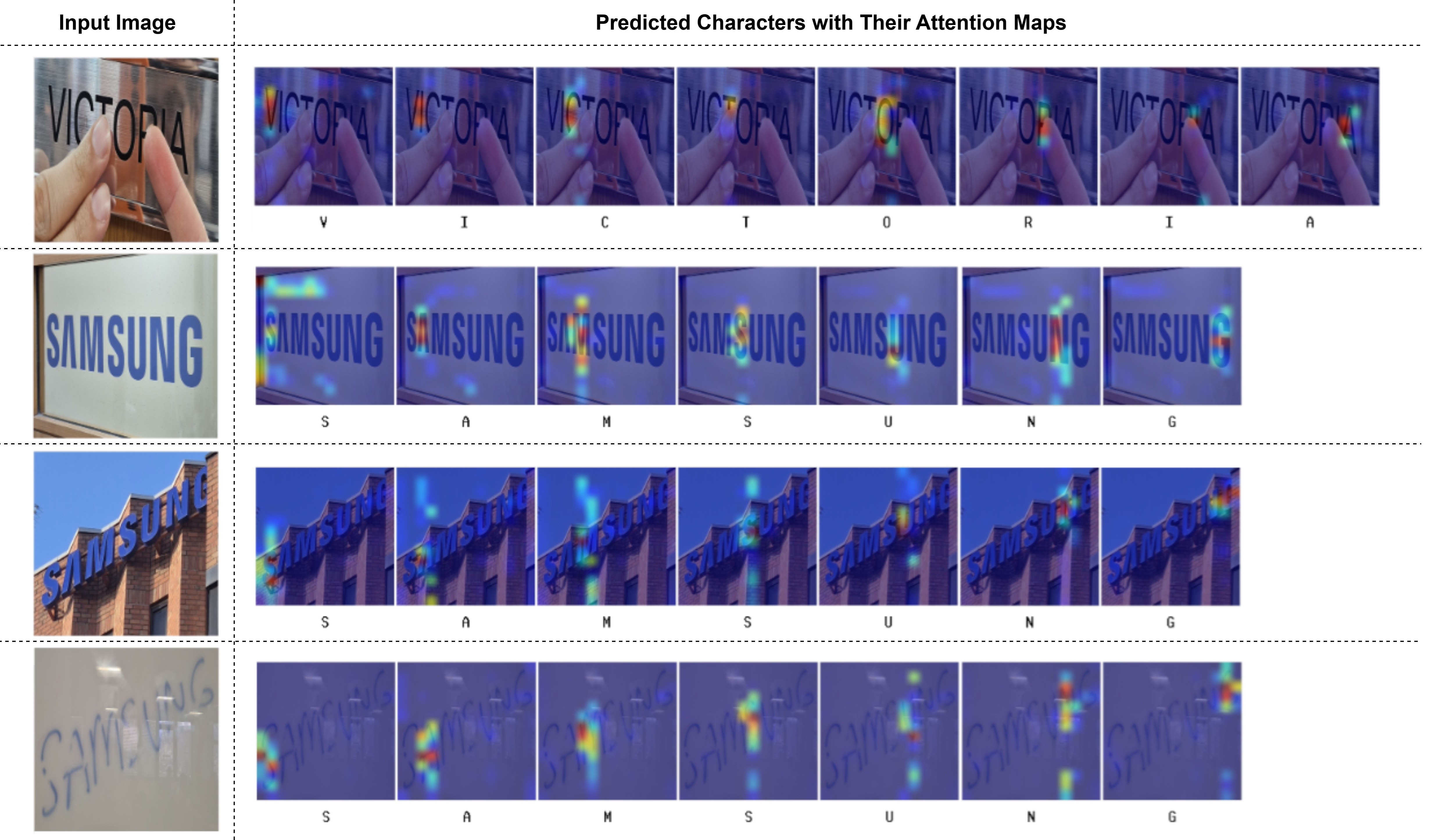}
\vspace{-0.8em}
\caption{\textbf{Prediction results of our STR system on real-world examples by visualizing character attention maps separately}. }
\vspace{-0.8em}
\label{fig:demo}
\end{figure*}

\subsubsection{Multi-Level Cascaded Transformers}

In this section, we explore the extension of cascaded transformers to multiple levels, where the number of cascaded models is set to $C>2$. This means that vision token size provided to the decoder also becomes smaller. The experimental results are reported in Table~\ref{tab:results_mult}. Notably, $e$-cc(3:3:3:3)-model$_{id}$ has the smallest computational complexity. The results show that the performance drop can be substantial for small-capacity transformers, such as \textit{tiny}. In contrast, for larger models, such as \textit{base}, the multi-level cascaded transformers yield comparable results, demonstrating their potential for efficient and accurate scene text recognition.

\begin{table}
  \centering
  
  %\settowidth\rotheadsize{\tb{ENCODER}}
  \addtolength{\tabcolsep}{-0.3em}
  \resizebox{0.8\linewidth}{!}{
  \begin{tabular}{c|cccc}
    \toprule
    & \multicolumn{2}{c}{$d$-tiny} & \multicolumn{2}{c}{$d$-small} \\
    & \textbf{Acc} & \textbf{Params} & \textbf{Acc} & \textbf{Params} \\
    \midrule
    $e$-tiny            & 91.01 & 15.1 M & 91.77 & 24.6 M \\
    $e$-cc(3:18)-tiny   & 91.29 & 19.1 M & 91.93 & 28.6 M \\
    $e$-cc(6:12)-tiny   & 91.22 & 17.8 M & 91.86 & 27.3 M \\
    $e$-cc(9:6)-tiny    & 91.41 & 16.5 M & 91.99 & 26.0 M \\
    \midrule
    $e$-small           & 92.24 & 31.3 M & 92.49 & 40.8 M \\
    $e$-cc(3:18)-small  & 92.62 & 47.2 M & 92.61 & 56.7 M \\
    $e$-cc(6:12)-small  & 92.42 & 41.9 M & 92.54 & 51.4 M \\
    $e$-cc(9:6)-small   & 92.56 & 36.6 M & 92.73 & 46.1 M \\
    \midrule
    $e$-base            & 92.77 & 95.4 M  & 93.21 & 104.9 M \\
    $e$-cc(3:18)-base   & 92.96 & 158.9 M & 93.32 & 168.1 M \\
    $e$-cc(6:12)-base   & 92.93 & 137.6 M & 93.23 & 147.1 M \\
    $e$-cc(9:6)-base    & 92.90 & 116.6 M & 93.26 & 126.1 M \\
    \bottomrule
  \end{tabular} }
  \caption{\textbf{Results for cascaded transformers with the same complexity of standard transformer.} Average word accuracy (\%) on the test benchmarks is reported with the parameter size of overall STR system.}
  \vspace{-0.8em}
  \label{tab:results_cmpl}
\end{table}

\subsubsection{Cascaded Transformers with the Same Complexity of Standard Transformer}

In this section, we specifically investigate the impact of cascaded-transformers when the model complexity is same to that of standard encoders, but with an increased representation capacity. Therefore, we adjust the computational complexity of cascaded-transformers by increasing the block size of transformers after vision token reduction. The results are reported in Table~\ref{tab:results_cmpl}. The results validate that cascaded-transformers improve the accuracy compared to the standard encoders.

\begin{table}
  \centering
  
  %\settowidth\rotheadsize{\tb{ENCODER}}
  \addtolength{\tabcolsep}{-0.3em}
  \resizebox{0.9\linewidth}{!}{
  \begin{tabular}{c|cccccc}
  \toprule
  &  & & & & \textbf{IC13} & \textbf{IC15} \\
  & \textbf{GFLOPs} & \textbf{ArT} & \textbf{COCO} & \textbf{HOST} & \textbf{1015} & \textbf{1811} \\
    \midrule
    CLIP4STR-B & 39.8  & 85.8 & 81.3 & 79.3 & 98.6 & 91.4 \\
    CLIP4STR-L & 171.9 & \textbf{86.4} & \textbf{82.7} & \underline{81.1} & \textbf{99.0} & \underline{91.9} \\
    \midrule
    $e$-cc(6:6)-base+$d$-small  & 29.9 & \underline{85.9} & \underline{82.4} & \textbf{83.1} & \underline{98.8} & \textbf{92.1} \\
    \bottomrule
  \end{tabular} }
  \caption{\textbf{Comparison with a SOTA baseline.} Word accuracy (\%) for different test benchmarks is reported with the GFLOPs of overall STR system.}
  \vspace{-0.8em}
  \label{tab:sota}
\end{table}

\subsubsection{Comparison with State-of-the-Art} We compare the accuracy and computational efficiency of our system with two variants of a SOTA model, CLIP4STR, which differ in model capacity. The results are summarized in Table~\ref{tab:sota}. They show that our system outperforms CLIP4STR-B model while requiring fewer computations. Moreover, our system obtains comparable results to CLIP4STR-L, but the computational requirement is reduced approximately 10x less.

\subsubsection{Remarks} There are several remarks in the experiments that we need to summarize: 

\begin{itemize}[leftmargin=*]
\item The key intution behind the negligible drop in accuracy with our cascaded model is that the most crucial layers of VIT for STR are concentrated in the early layers. As a result, progressively decreasing the token sizes has a minimal impact on accuracy. Notably, similar observations have been made in language models~\cite{lan2019albert, men2024shortgpt}, suggesting that the early layers of transformers are more important to extract robust representations and impact the accuracy more than other layers.

\item Our cascaded model is able to learn which tokens are important and which are redundant, allowing it to adaptively focus on the most relevant information.
    
\item Given that our model maintains the same number of layers and representation dimensionality, the memory requirements remain unchanged. Specifically, we conduct an analysis of various encoder and decoder configurations, each can be tailored to specific device requirements. In comparison to our model, which has approximately 100M parameters, a base model, CLIP4STR, has significantly larger parameter sizes, with 160M parameters for its base version and 450M parameters for its large version.
    
\end{itemize}

\subsubsection{Results on Real-World Examples}

We implement a custom software setup in Python, running on a desktop computer, to test our STR system on real-world images. In our system, a user can copy images from the web or a mobile phone into a folder, and then select the images to run the STR algorithm\footnote{Video link: https://drive.google.com/drive/folders/1OPFWi68iK0gb7EC74Yh9t3S7euwbPawI}. The outputs of character predictions with separate attention maps for each predicted character are visualized in Figure~\ref{fig:demo}. Notably, test images are deliberately selected to be highly distorted, occluded and rotated, allowing us to demonstrate the robustness of our STR system for challenging conditions.

\section{Conclusion}

In this work, we present an efficient and accurate scene text recognition (STR) system. To improve the efficiency, we introduce a cascaded-transformers structure, which comprises multiple sub-transformers that selectively retain a subset of vision tokens at each level. Eventually, our system decreases the computational requirements by eliminating the redundancy between vision tokens. Experimental results demonstrate that with only a slight accuracy drop, our system substantially improves the efficiency of STR. Furthermore, we evaluate our STR system on real-world images using a custom software setup. We visualize that our system is robust to various adverse conditions such distortion, occlusion and rotation.

\bibliographystyle{ACM-Reference-Format}
\bibliography{main}

\end{document}